\documentclass[conference]{IEEEtran}
\ifCLASSINFOpdf
\else
\fi

\usepackage{graphicx}
\usepackage{graphics}

\usepackage{subcaption}

\begin{document}
%
\title{Fruit and Vegetable Identification Using Machine Learning for Retail Applications}


\author{\IEEEauthorblockN{Frida Femling, Adam Olsson, Fernando Alonso-Fernandez}
\IEEEauthorblockA{\textit{School of Information Technology} \\
Halmstad University, SE 301 18, Halmstad, Sweden \\
frifem15@student.hh.se, adaols15@student.hh.se, feralo@hh.se}
}


%


\maketitle

\begin{abstract}
This paper describes an approach of creating a system identifying fruit and vegetables in the retail market using images captured with a video camera attached to the system. The system helps the customers to label desired fruits and vegetables with a price according to its weight. The purpose of the system is to minimize the number of human computer interactions, speed up the identification process and improve the usability of the graphical user interface compared to existing manual systems.
The hardware of the system is constituted by a Raspberry Pi, camera, display, load cell and a case.
To classify an object, different convolutional neural networks have been tested and retrained.
To test the usability, a heuristic evaluation has been performed with several users,
concluding that the implemented system is more user friendly compared to existing systems.
\end{abstract}


%
\IEEEpeerreviewmaketitle

\section{Introduction}

Complex and time consuming self-service systems may result in customers choosing another grocery store. Since customers are the reason companies survive, their satisfaction is the businesses’ key to success. The necessity of systems which decreases the process time exists because of consumers’ expectations of their constant endeavor to save time.

Accordingly, the purpose of this work is to improve the identification process of fruit and vegetables performed by the self-service systems in the retail market. More specifically, the improvement should consist of a faster process and a more user friendly system. The purpose of implementing computer vision to the system is to narrow the selection of possible objects and thus reduce the strain on  the user. Additionally, the use of computer vision in self-service systems can simplify the process of identifying objects by moving the process from a human to a computer. Theoretically, this could hasten the process to identify products and minimize the amount of errors by removing the human factor.

\begin{figure} [b]
\centering
\includegraphics[width=0.42\textwidth]{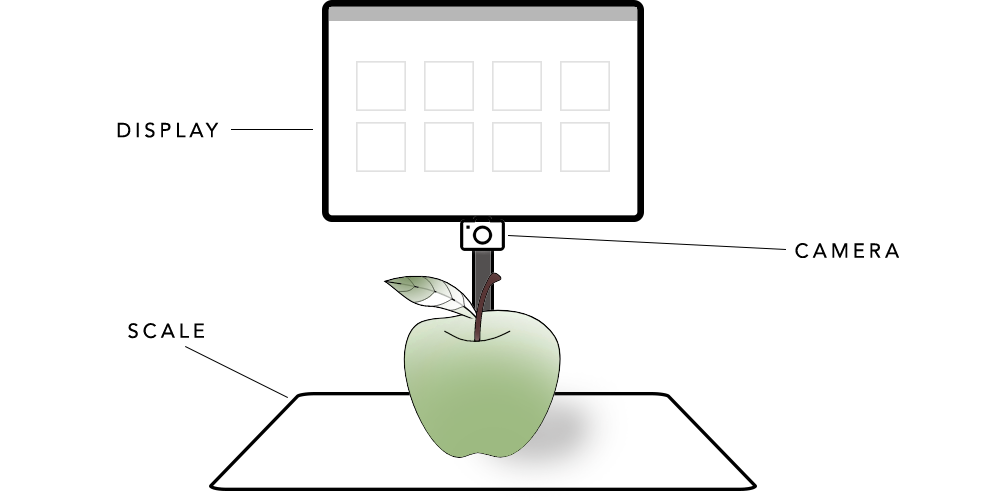}
\caption{Prototype of our system.}
\label{fig:prototype}
\end{figure}

The work of this paper will be limited to the fundamentals of the identification system.  In terms of hardware, the fundamentals are a camera, a display, an activation mechanism representing the scale, and a processor to run the system (Figure~\ref{fig:prototype}). An image classifier has been trained and evaluated to classify images of fruits and vegetables from the camera. For classification purposes, we will investigate convolutional neural networks (CNN) architectures, given the huge success shown in recent years by CNNs in several object recognition and classification tasks \cite{[Lecun15],[Schmidhuber15]}. A user interface has been developed as well to handle user interaction via the display. The display will show a graphical user interface (GUI) for the user to interact with and present the classifiers output. The system will exclude a label printer, but simulate the process in the graphical user interface.

\subsection{Related Works}
Identification of fruits and vegetables are implemented in different areas. The most common areas are identification in the retail business, and in areas where the purpose is to ease the harvest in the perspective of agriculture. In the retail business, the identification is mostly done manually by a cashier, or via the self-service systems in a store.

A company which has made great progress in its technical evolution when it comes to artificial intelligence, image recognition and automating physical work is Amazon\footnote{ttps://www.amazon.com/}. Amazon developed a product, called Amazon Go, which enabled a shopping experience without cashiers or self-service checkouts. The company built the store where the customers check in with a smart phone using the application Amazon Pay. The store is set up with a large amount of cameras and sensors. Thanks to computer vision and deep learning algorithms, Amazon managed to create a store where technology identifies the products the customers choose. No checkout is required, the chosen products are debited from the Amazon Pay account of the costumer.

StrongPoint\footnote{https://www.strongpoint.se/} is a company, with its headquarters in Norway, offering technical solutions to the retail business. StrongPoint recently released an identification system called Digi. Digi consists of a user interface displayed on a touchscreen, a scale, a camera and a label printer. The software is implemented with image recognition in the identification process and can be compared to the existing counterpart of this project. Digi is new to the market, hence it is not used in many stores.

Related work including image recognition has been done in the purpose of controlling the vegetation and harvest of fruit and other growths at fields of farmers \cite{[Bargoti17],[Sa16],[Nuske11],[Yamamoto14]}. The technology has been used to automate the yield with the help of robotic harvesting. Several CNNs has been used to localize the fruits in the purpose of either collecting or counting. However, the issue of creating a fast and reliable fruit detection system persists \cite{[Kapach12]}. This is due to  large variation in the appearance of the fruits in field, including colour, shape, size  and  texture properties.

\subsection{Heuristic Evaluation}
Nowadays, the identification of products in the retail business is done manually, either by the cashier or the consumer with the help of self-service systems. When the identification is performed manually, the human factor may affect the outcome. There is a chance that the user press the wrong button or could misinterpret the application. An evaluation of existing systems or new developments could help to create better GUI that can replace existing solutions for more efficient ones.

Heuristic evaluation is an informal method of usability analyses for user interfaces \cite{[Nielsen90],[Nielsen92]}. It is simply done by looking at an interface to gather an opinion of what is positive and negative about the interface. Formal collections of guidelines exist when developing interfaces but may come across as intimidating since they are in an order of thousands \cite{[Molich90],[Smith86]}. These guidelines has been reduced to nine guidelines that capture the most crucial errors \cite{[Molich90]}:

\begin{itemize}
\item Simple and Natural Dialogue
\item Speak the User’s Language
\item Minimize the User’s Memory Load
\item Be consistent
\item Provide Feedback
\item Provide Clearly Marked Exits
\item Provide Shortcuts
\item Provide Good Error Messages
\item Error Prevention
\end{itemize}

Experiments found that a single participant following these nine guidelines are rarely able to find more than 50\% of usability problems \cite{[Nielsen90]}.  However, by aggregating multiple problems identified by the participants, the heuristic evaluation method performs quite well. Only as few as three to five participants are in most cases able to find more than 70\% of the usability problems \cite{[Nielsen90]}.

\section{Methodology}

The project is divided into two phases, the experimentation phase and the implementation phase. The experimentation phase aims to find the most suitable network for this project. It lays the foundation of how the system will perform in the end. The implementation phase aims to describe how the software and hardware are integrated to form an identification system.

\subsection{Convolutional  Neural  Networks}

Convolutional  Neural  Networks (CNN) have over the recent years become great at large-scale image recognition tasks. Large-scale image recognition has been become possible because of large public image databases such as ImageNet. In this paper, we employ transfer learning by selecting some pre-trained architectures, and fine-tuning them to the type of images used in our application \cite{[Pan10]}. In many real-world applications, it is expensive or impossible to recollect the needed training data and rebuild the models from scratch. Transfer learning is thus a way to create new models with very little data compared to the initial training. Network architectures and open source scripts for retraining are provided by Tensorflow.

The number of available architectures to train goes beyond count. Comparing all architectures to each other is a difficult task. Instead, the Inception and MobileNet architectures has properties making them worth while evaluating for this project.
Inception v3 is an open source architecture created by Google and trained  on 1.2 million images from thousands of different categories. It is a module of GoogleLeNet designed to function under strict constraints on memory and on a computational budget \cite{[Szegedy16]}. In the ImageNet Challenge 2014, GoogleLeNet with the Inception v3 module, had the least error rate comparing to other architectures \cite{[Russakovsky15]}. With an average error rate of 6.66\%, the network defeated all the other competitors. The Inception v3 module is 42 layers deep.
In many real time mobile applications implemented with recognition tasks to identify certain objects or surroundings, light weight architectures are preferable to match the resource restrictions on the platforms. MobileNet is an architecture developed to function on mobile and embedded vision applications \cite{[Howard17corr]}. MobileNet is built on depthwise separated convolutions to reduce the computation and model size. The depthwise separated convolutions splits the standard convolution method of combining and filtering into different layers, one layer for combining and one layer for filtering. This method reduces the computation size drastically. Almost 75\% of the total parameters in the network are located in convolutional layers using a kernel of 1$\times$1, which is what reduces  the computation size. Depthwise, MobileNet has 28 layers.

\begin{figure} [t]
\centering
\includegraphics[width=0.42\textwidth]{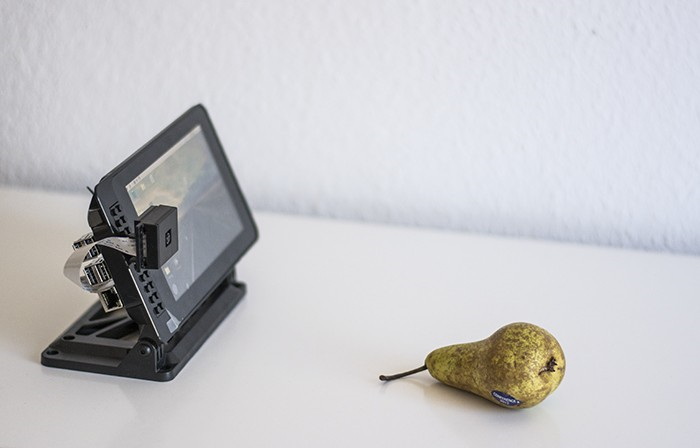}
\caption{Setup of our prototype with the hardware employed.}
\label{fig:hardware}
\end{figure}

\begin{figure*} [t]

\centering
\includegraphics[width=0.96\textwidth]{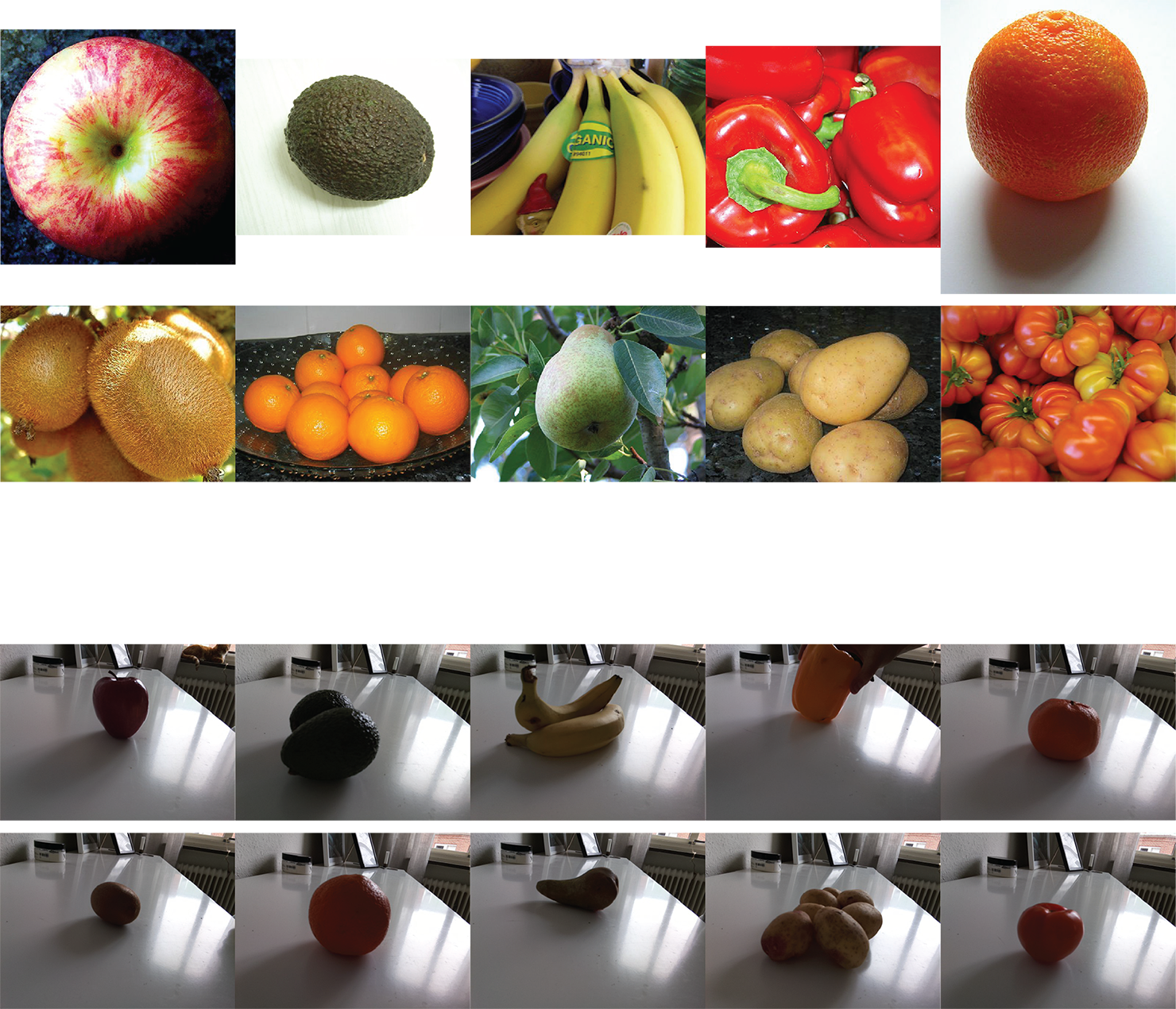}
\caption{Example images from each class. Top row: images from ImageNet. Bottom row: self-collected images.}
\label{fig:data}
\end{figure*}

\subsection{Hardware}

To replicate a real life scenario where a self-serving system has limited processing power and physical space, a Raspberry Pi has been selected. A Raspberry Pi operates in a way similar to a regular computer but on a fraction of the cost and size. It is  a great development platform for creating prototypes and trying concepts. It is based on a 64bit Quad Core 1.2GHz CPU and has 1GB of RAM available.  Additionally, the Raspberry Pi has a CSI and a DSI port for connecting a Raspberry Pi camera module and a touchscreen display. Additionally, the Raspberry Pi also has 40 general purpose pins to connect various hardware.  The processor has a wireless LAN that enables an Internet connection. Furthermore, CNN’s and various deep learning frameworks has been benchmarked to use on the mini computer \cite{[Pena17]}.

The camera used for this project is the Raspberry Pi Camera Module v2 and  is the official product from the Raspberry Pi Foundation.  The camera has
8-megapixels resolution and is compatible with the Raspberry Pi without any drivers which enables a quick setup. The camera is connected via a ribbon cable to the DSI port on the Raspberry Pi.

The touchscreen display used for this project is called Raspberry Pi Display  7” multitouch. The display is a 800 by 480 pixels display and connects to the processor via an adapter board. The adapter board handles power and signal conversion to the Raspberry Pi. The display is connected to the GPIO port to get power from the processor and a second connection is required to the DSI port to visualize the  data.

\begin{figure*} [t]
\centering
    \begin{subfigure}{0.45\linewidth}
		\includegraphics[width=\linewidth]{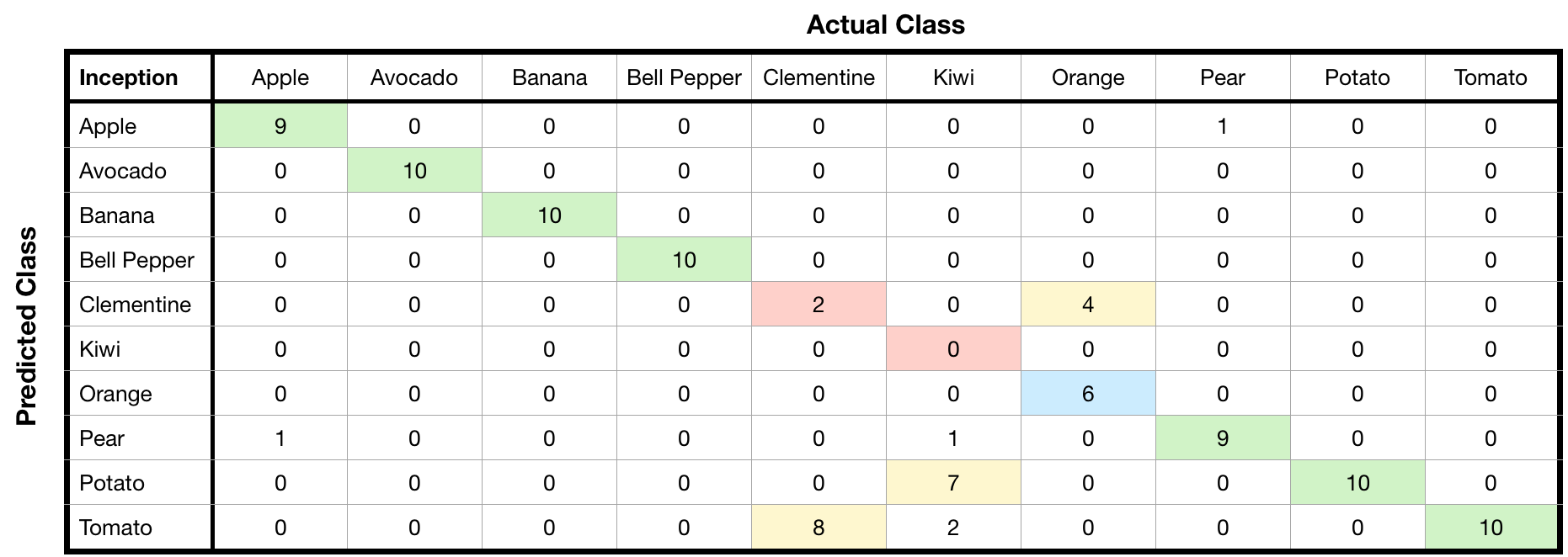}
        \caption{Inception}
	\end{subfigure}
    \begin{subfigure}{0.45\linewidth}
		\includegraphics[width=\linewidth]{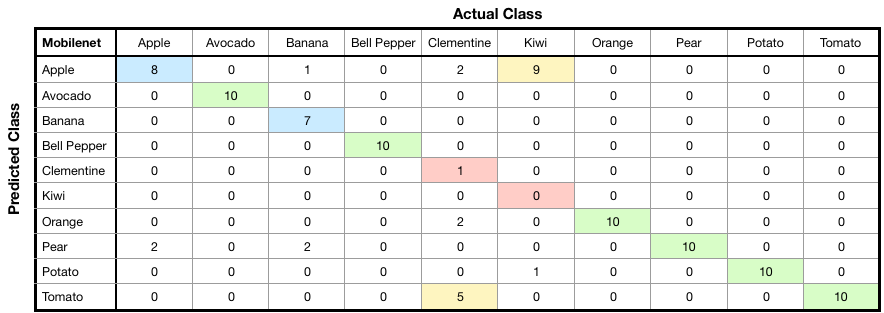}
        \caption{MobileNet}
	\end{subfigure}
\caption{Classification results: confusion matrix (top 1).}
\label{fig:top1}
\end{figure*}

\begin{figure*} [t]

\centering
    \begin{subfigure}{0.45\linewidth}
		\includegraphics[width=\linewidth]{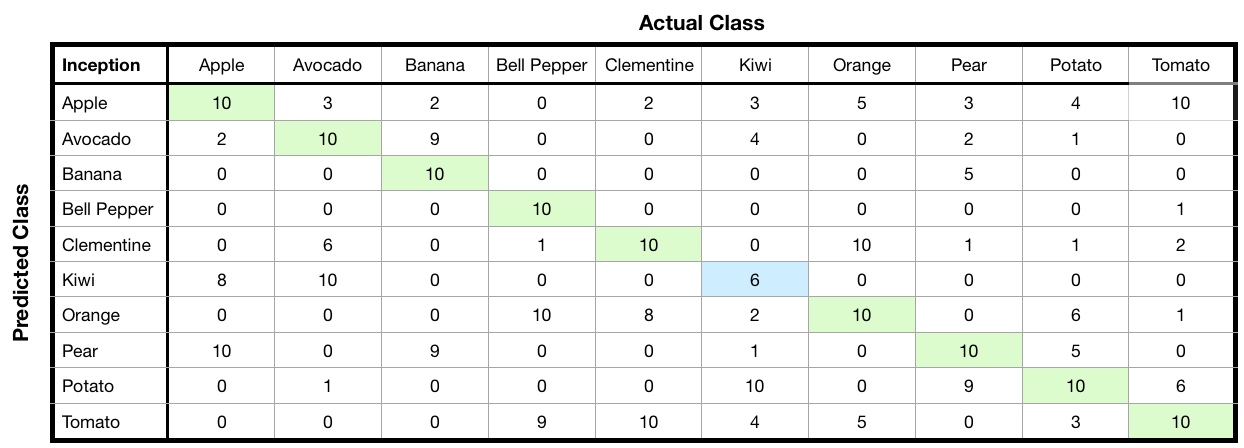}
        \caption{Inception}
	\end{subfigure}
    \begin{subfigure}{0.45\linewidth}
		\includegraphics[width=\linewidth]{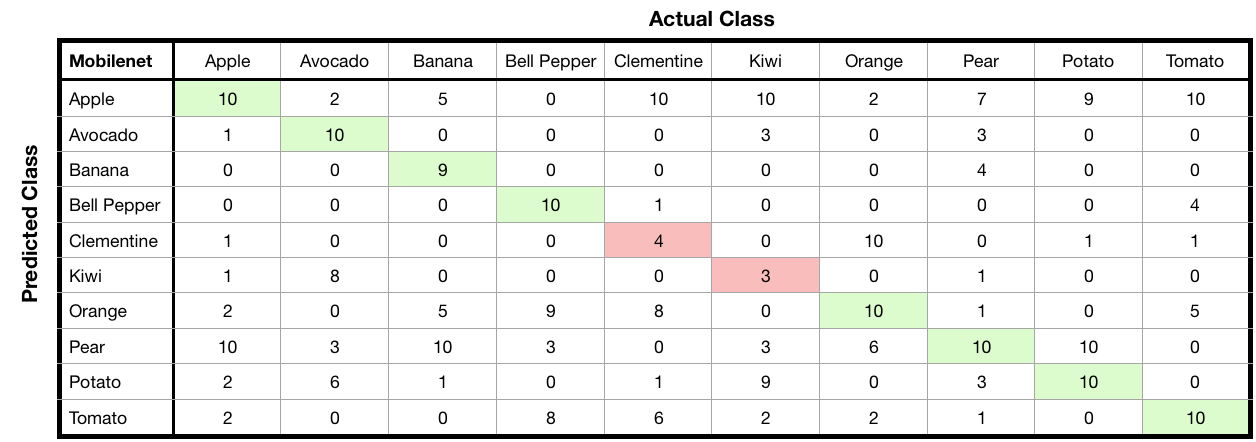}
        \caption{MobileNet}
	\end{subfigure}
\caption{Classification results: confusion matrix (top 3).}
\label{fig:top3}
\end{figure*}

\subsection{Dataset}

We employ 10 different classes in our study. The chosen classes are apple, avocado, banana, bell pepper, clementine, kiwi, orange, pear, potato and tomato. These classes are chosen because some fruits and vegetables have similar appearances and are frequently bought in retail markets. Limitations to the data set has been done in order to not make the project to extensive. These limitations are that all types of a fruit or vegetables reside under the same class. This means all types of apples reside under the apple class and similar for each fruit.

A dataset with 400 images per class has been extracted from ImageNet. 
In addition, 30 images per class have also been collected with the camera employed in this project. For simplicity, images of fruit and vegetables has been taken without being placed in plastic bags.
Example images from each class can be found in Figure \ref{fig:data}.

\section{Experiments and Results}

\subsection{Image Classification}

The Convolutional  Neural  Networks are evaluated by two properties: propagation time, which is the time it takes for an image to be classified, and accuracy, which is how accurate the prediction is.

To evaluate the propagation time of the networks, series of 100 images was captured and classified. The propagation time is the time between start and end of the classification. Each test of a network was run five times and yielded 500 samples of time propagation per network. Additionally, only one network was loaded per test to prevent filling the working memory and affecting outcome of the propagation time. Between the two networks there is a large difference in propagation time. The average propagation times for 500 images are 3.3 seconds for Inception and
0.43 seconds for MobileNet.
This makes the average propagation time of Inception 7.67 times slower than MobileNet.
When measuring propagation times, the first image of each test for both networks had a significantly longer propagation time. This is speculated, but not confirmed, to be because the application is not loaded into the cache memory. Once the first image has propagated, the network has all the operations loaded into cache and thus can access them faster.

\begin{figure*} [t]
\centering
\includegraphics[width=0.3\textwidth]{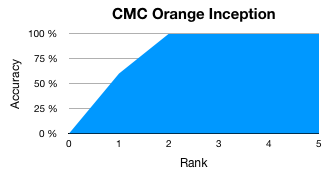}
\includegraphics[width=0.3\textwidth]{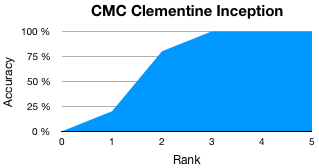}
\includegraphics[width=0.3\textwidth]{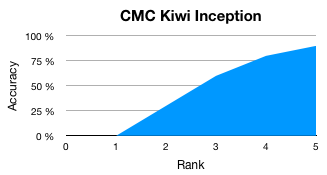}
\caption{CMC curves of selected classes with Inception.}
\label{fig:inceptionCMC}
\end{figure*}

\begin{figure*} [t]
\centering
\includegraphics[width=0.3\textwidth]{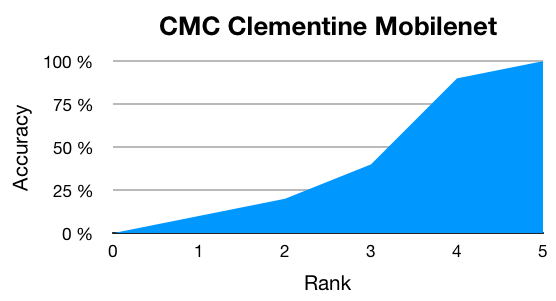}
\includegraphics[width=0.3\textwidth]{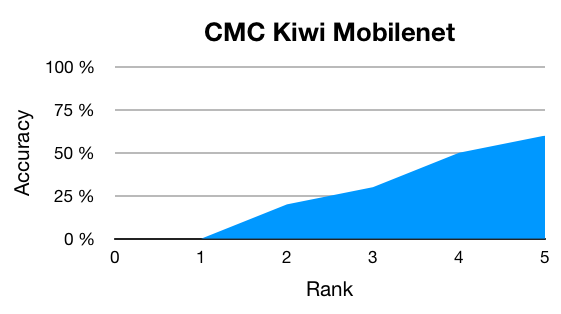}
\caption{CMC curves of selected classes with MobileNet.}
\label{fig:mobilenetCMC}
\end{figure*}

Regarding accuracy tests,
images of fruits and vegetables captured by the mounted camera was propagated through the network, and the result of each classification was logged to a file. 
A total of 10 tests was conducted on each network. Each test yielded 10 samples of accuracy on fruits or vegetables from different angles and in various amounts. This resulted in 100 samples of accuracy for each  network.
The confusion matrices of Inception and MobileNet are shown in Figure \ref{fig:top1}. A network top 1 accuracy can be calculated by dividing the sum of the diagonal by the number of tests performed.
Three types of color markings are used. Green, which are classes where the network performs with accuracy higher than 90\%. Blue, classes that are predicted to be correct more than half of the time but not viable enough to get a green marking. Red, classes the network fails to predict. In addition, the confusion matrices of Inception and MobileNet for the top 3 predictions are shown in Figure \ref{fig:top3}, i.e. the number of tests where the correct class is output in the first three positions.
The top 1 accuracy of Inception in 76\%. Highlighted in green are apple, avocado, banana, pear, potato and tomato. Marked in blue is orange which is often misinterpreted as a clementine. Finally, marked in red are clementine and kiwi. Inception has a problem distinguishing an orange from an clementine, but an actual clementine is often misinterpreted as a tomato.
Figure \ref{fig:inceptionCMC} shows the CMC for the orange, clementine, and kiwi classes. It shows that even though orange was misinterpreted as a clementine, the correct label was always among the top 2 ranks during the test.
A 100\% percent with clementines is reached when displaying top 3 rankings of each prediction.
Kiwi is by far the most difficult class for Inception to label. Not even at a top 5 ranking is the network able to have a 100\% accuracy for kiwis.
The overall top 3 accuracy of Inception is 96\%.

The performance of MobileNet, in terms of accuracy, is quite similar to the performance of Inception. The top 1 accuracy is the same as for Inception, 76\%. Marked in green are avocado, bell pepper, orange, pear, potato and tomato. Marked in blue are apple and banana. These two classes still perform quite well, even though marked in blue. Apple is misinterpreted for a pear  two times and banana is misinterpreted three times, one of the times as an apple and twice for a pear. The classes MobileNet has difficulties interpreting are clementine and kiwi. When comparing to Inception, MobileNet is actually better at classifying oranges and pears. However, MobileNet performs worse or equally good in all other categories.
A look at the CMC curves of MobileNet (Figure \ref{fig:mobilenetCMC}) shows that it reaches a 100\% accuracy for clementine if the top 5 labels are displayed. The network misinterpreted the clementine a majority of the times for a tomato.
Just as Inception has difficulties labeling kiwi, MobileNet follows in the same lines but for the worse. Out of 10 images, MobileNet succeeds in a 60\% accuracy when taking the top 5 ranks into consideration.
The overall top 3 accuracy of MobileNet is 97\%.

\begin{figure*} [t]
\centering
    \begin{subfigure}{0.45\linewidth}
		\includegraphics[width=\linewidth]{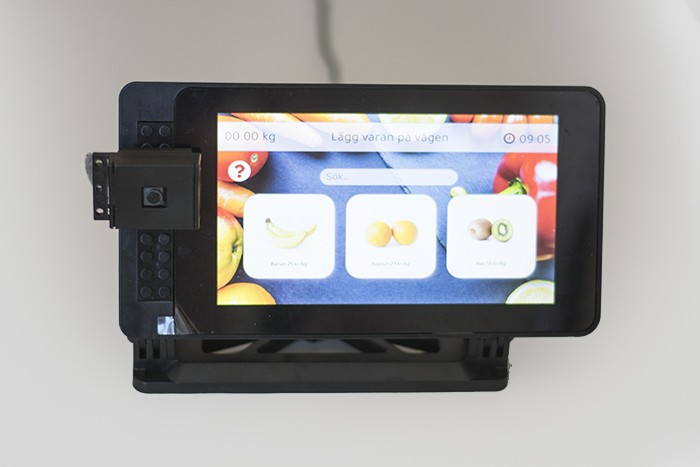}
        \caption{Front of the prototype displaying the GUI}
	\end{subfigure}
    \begin{subfigure}{0.45\linewidth}
		\includegraphics[width=\linewidth]{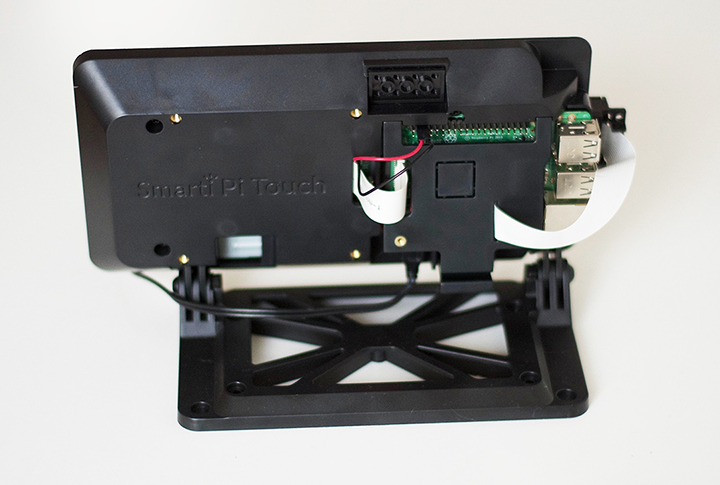}
        \caption{Back of the prototype}
	\end{subfigure}
    \begin{subfigure}{0.45\linewidth}
		\includegraphics[width=\linewidth]{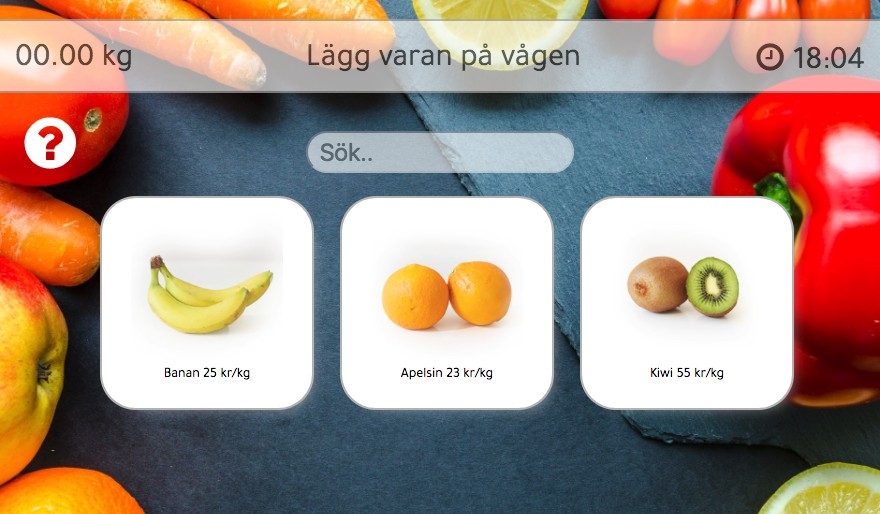}
        \caption{Default page of GUI}
	\end{subfigure}
    \begin{subfigure}{0.45\linewidth}
		\includegraphics[width=\linewidth]{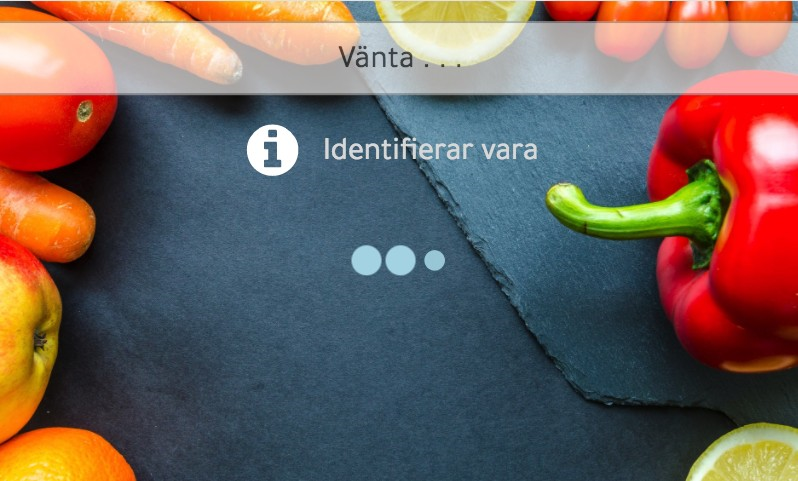}
        \caption{Identifying object}
	\end{subfigure}
\caption{Prototype showing the Graphical User Interface.}
\label{fig:GUI}
\end{figure*}

\subsection{Usability}

The design of the Graphical User Interface (GUI) is created with the goal of appearing simple and conspicuous (Figure \ref{fig:GUI}). It is supposed to lead the user in the right direction with few interaction calls. The user can either put the product on the activation mechanism which triggers the classifier, search for the product or chose the product directly if it is displayed at the default page. The products displayed at the default page is the most frequently bought products and are chosen manually. Heuristic evaluations has been performed on the developed system. The evaluations concluded which of the nine guidelines the graphical user interface breaks.

An initial Prelaunch Test concluded that there was some usability flaws in the graphical user interface, which helped us to refine the design. The most frequently usability flaw occurred when the user was supposed to print the label. When the system finished to identify the fruit or vegetable and the result was presented on the display, some of the users did not click the product to print the label. The users assumed that the label was going to print automatically when the correct fruit or vegetable was identified. However, the users who got two or more products as a result, did click the product.
According to the Prelaunch Test, the discovered usability flaws was \cite{[Molich90]}:

\begin{itemize}
\item No clear printing request
\item No clear feedback
\end{itemize}

The test users provided some valuable input such as:
\begin{itemize}
\item Add an instruction button
\item The header-component looks like its clickable because of rounded corners
\end{itemize}

According to the subsequent heuristic evaluation on five test persons \cite{[Nielsen90]}, the resulted system broke three of the guidelines:

\begin{itemize}
\item Speak the User’s language
\item Provide feedback
\item Provide clearly marked exits
\end{itemize}

The most frequently broken guideline was \textit{Provide Feedback}. Some of the users wanted clearer instructions about what process was running.  Several users
pointed out that they expected a clearer message of when the identification was performed.
Another guideline which was frequently broken was \textit{Provide Clearly Marked Exits}. This occurred because the user was supposed to click the product to get a label, since the system did not provided any indication that the identification is finished.
Regarding \textit{Speak the User’s language}, some users did not like the background of the display, or suggested to use a bigger display.

After the indicated flaws were fixed, another heuristic evaluation was performed on five test persons. Three of the volunteers had not tested the system before. According to the result, no guidelines were broken and all test persons completed the identification within seconds.

We also measured the time of identifying a fruit or vegetable using the developed system. The time is measured from when the user puts the product on the scale to when the label is printing. This test is perfomed on 10 users. An average time of 10.1 seconds was measured, with a minimum of 5.77 seconds (fastest user), and a maximum of 20 seconds.

\section{Discussion}

This paper has presented a system that makes use of computer vision to automatize the identification process of fruits and vegetables by self-service systems in the retail market. Another goal has been to create a user friendly system, as measured by usability studies.
We have evaluated two Convolutional Neural Network architectures (Inception and MobileNet) \cite{[Szegedy16],[Howard17corr]} as classifiers of 10 different kinds of fruits or vegetables. Images for the classifier are provided by a Raspberry Pi Camera Module v2, connected to a Raspberry Pi. The system is complemented with a touch screen display, and a graphical user interface which presents the detected classes to the user.

MobileNet provided fast identification results with accurate predictions. However, the differences in accuracy between the networks is not as large as the difference between  the propagation times. MobileNet propagates images significantly faster with almost the same accuracy. MobileNet has a top 3 accuracy of 97\%. However, even though the top 3 accuracy is great, MobileNet still has difficulties in predicting clementines and kiwis.
As for kiwi, the accuracy could be improved by creating a new set of images for this class. Currently, a majority of the images depicts kiwis being cut open  and showing their green insides. These images does not depict what the kiwi looks like in the working environment and the network will thus not recognize the fruit.
Additional input such as weight could also be taken into consideration to better differentiate these two fruits, although this may demand to count the number of pieces presented to the scale.

Performing retraining on data sets from its actual environment could get the network more accurate \cite{[Pan10]}. Using data sets fetched from ImageNet \cite{[Russakovsky15]} resulted in pictures of fruits and vegetables in varying environments. Since the classifier is supposed to work in a retail store, it will never encounter images of for example a forest. Therefore, training images that differ from the working environment could probably confuse the classifier. Furthermore, if more images are collected, classes can be split into subclasses containing different types of a fruit or vegetable. For example, the apple class could be split into subclasses like Granny Smith, Pink Lady and Royal Gala which are all types of apples. However,  there is a risk that splitting a class into subclasses for each type    of fruit or vegetable turns out to be too challenging for the network to classify. A more accurate behaviour may be achieved by implementing a series of networks. The first network only decides what kind of fruit it is. The task of classifying the subset of fruit is left to one of many succeeding networks specialized on a single kind of fruit.

The heuristic evaluation of the developed system \cite{[Molich90],[Nielsen90]} indicated a positive result.
Our system lacked some functions and the users had some valuable inputs which helped us to improve the design.
After three iterations, users indicated that there was no usability flaws.
According to the usability tests, a bigger display was desirable. A consequence of a small display limited the font size and the product item size. However, a larger display may affect the processing time.

To get this product into business, several concerns needs to be taken into consideration. Since this  product includes a camera to take photos, there might be a risk of catching a persons’ face in the picture. A possible solution is to place the camera looking downwards from above the scale, thus minimizing this effect.

\section*{Acknowledgments}

Author F. A.-F. thanks the Swedish Research Council (VR), the Swedish Knowledge Foundation (KK) and the Swedish Innovation Agency (VINNOVA) for funding his research.



%

\bibliographystyle{IEEEtran}

\begin{thebibliography}{10}
\providecommand{\url}[1]{#1}
\csname url@samestyle\endcsname
\providecommand{\newblock}{\relax}
\providecommand{\bibinfo}[2]{#2}
\providecommand{\BIBentrySTDinterwordspacing}{\spaceskip=0pt\relax}
\providecommand{\BIBentryALTinterwordstretchfactor}{4}
\providecommand{\BIBentryALTinterwordspacing}{\spaceskip=\fontdimen2\font plus
\BIBentryALTinterwordstretchfactor\fontdimen3\font minus
  \fontdimen4\font\relax}
\providecommand{\BIBforeignlanguage}[2]{{%
\expandafter\ifx\csname l@#1\endcsname\relax
\typeout{** WARNING: IEEEtran.bst: No hyphenation pattern has been}%
\typeout{** loaded for the language `#1'. Using the pattern for}%
\typeout{** the default language instead.}%
\else
\language=\csname l@#1\endcsname
\fi
#2}}
\providecommand{\BIBdecl}{\relax}
\BIBdecl

\bibitem{[Lecun15]}
Y.~Lecun, Y.~Bengio, and G.~Hinton, ``\BIBforeignlanguage{English (US)}{Deep
  learning},'' \emph{\BIBforeignlanguage{English (US)}{Nature}}, vol. 521, no.
  7553, pp. 436--444, 5 2015.

\bibitem{[Schmidhuber15]}
\BIBentryALTinterwordspacing
J.~Schmidhuber, ``Deep learning in neural networks: An overview,'' \emph{Neural
  Networks}, vol.~61, pp. 85 -- 117, 2015. [Online]. Available:
  \url{http://www.sciencedirect.com/science/article/pii/S0893608014002135}
\BIBentrySTDinterwordspacing

\bibitem{[Bargoti17]}
S.~Bargoti and J.~Underwood, ``Deep fruit detection in orchards,'' in
  \emph{IEEE International Conference on Robotics and Automation (ICRA)}, May
  2017, pp. 3626--3633.

\bibitem{[Sa16]}
I.~Sa, Z.~Ge, F.~Dayoub, B.~Upcroft, T.~Perez, and C.~McCool, ``Deepfruits: A
  fruit detection system using deep neural networks,'' \emph{Sensors}, vol.~16,
  no.~8, 2016.

\bibitem{[Nuske11]}
S.~Nuske, S.~Achar, T.~Bates, S.~Narasimhan, and S.~Singh, ``Yield estimation
  in vineyards by visual grape detection,'' in \emph{IEEE/RSJ International
  Conference on Intelligent Robots and Systems}, Sept 2011, pp. 2352--2358.

\bibitem{[Yamamoto14]}
\BIBentryALTinterwordspacing
K.~Yamamoto, W.~Guo, Y.~Yoshioka, and S.~Ninomiya, ``On plant detection of
  intact tomato fruits using image analysis and machine learning methods,''
  \emph{Sensors}, vol.~14, no.~7, pp. 12\,191--12\,206, 2014. [Online].
  Available: \url{http://www.mdpi.com/1424-8220/14/7/12191}
\BIBentrySTDinterwordspacing

\bibitem{[Kapach12]}
\BIBentryALTinterwordspacing
K.~Kapach, E.~Barnea, R.~Mairon, Y.~Edan, and O.~Ben-Shahar, ``Computer vision
  for fruit harvesting robots \&\#150; state of the art and challenges ahead,''
  \emph{Int. J. Comput. Vision Robot.}, vol.~3, no. 1/2, pp. 4--34, Apr. 2012.
  [Online]. Available: \url{http://dx.doi.org/10.1504/IJCVR.2012.046419}
\BIBentrySTDinterwordspacing

\bibitem{[Nielsen90]}
\BIBentryALTinterwordspacing
J.~Nielsen and R.~Molich, ``Heuristic evaluation of user interfaces,'' in
  \emph{Proceedings of the SIGCHI Conference on Human Factors in Computing
  Systems}, ser. CHI '90.\hskip 1em plus 0.5em minus 0.4em\relax New York, NY,
  USA: ACM, 1990, pp. 249--256. [Online]. Available:
  \url{http://doi.acm.org/10.1145/97243.97281}
\BIBentrySTDinterwordspacing

\bibitem{[Nielsen92]}
\BIBentryALTinterwordspacing
J.~Nielsen, ``Finding usability problems through heuristic evaluation,'' in
  \emph{Proceedings of the SIGCHI Conference on Human Factors in Computing
  Systems}, ser. CHI '92.\hskip 1em plus 0.5em minus 0.4em\relax New York, NY,
  USA: ACM, 1992, pp. 373--380. [Online]. Available:
  \url{http://doi.acm.org/10.1145/142750.142834}
\BIBentrySTDinterwordspacing

\bibitem{[Molich90]}
\BIBentryALTinterwordspacing
R.~Molich and J.~Nielsen, ``Improving a human-computer dialogue,''
  \emph{Commun. ACM}, vol.~33, no.~3, pp. 338--348, Mar. 1990. [Online].
  Available: \url{http://doi.acm.org/10.1145/77481.77486}
\BIBentrySTDinterwordspacing

\bibitem{[Smith86]}
S.~L. Smith and J.~N. Mosier, ``Guidelines for designing user interface
  software,'' \emph{Technical Report The MITRE Corporation -
  http://www.dtic.mil/docs/citations/ADA177198}, 1986.

\bibitem{[Pan10]}
S.~J. Pan and Q.~Yang, ``A survey on transfer learning,'' \emph{IEEE
  Transactions on Knowledge and Data Engineering}, vol.~22, no.~10, pp.
  1345--1359, Oct 2010.

\bibitem{[Szegedy16]}
C.~Szegedy, V.~Vanhoucke, S.~Ioffe, J.~Shlens, and Z.~Wojna, ``Rethinking the
  inception architecture for computer vision,'' in \emph{Proc IEEE Conference
  on Computer Vision and Pattern Recognition, CVPR}, June 2016, pp. 2818--2826.

\bibitem{[Russakovsky15]}
O.~Russakovsky, J.~Deng, H.~Su, J.~Krause, S.~Satheesh, S.~Ma, Z.~Huang,
  A.~Karpathy, A.~Khosla, M.~Bernstein, A.~C. Berg, and L.~Fei-Fei, ``Imagenet
  large scale visual recognition challenge,'' \emph{International Journal of
  Computer Vision}, vol. 115, no.~3, pp. 211--252, Dec 2015.

\bibitem{[Howard17corr]}
\BIBentryALTinterwordspacing
A.~G. Howard, M.~Zhu, B.~Chen, D.~Kalenichenko, W.~Wang, T.~Weyand,
  M.~Andreetto, and H.~Adam, ``Mobilenets: Efficient convolutional neural
  networks for mobile vision applications,'' \emph{CoRR}, vol. abs/1704.04861,
  2017. [Online]. Available: \url{http://arxiv.org/abs/1704.04861}
\BIBentrySTDinterwordspacing

\bibitem{[Pena17]}
D.~Pena, A.~Forembski, X.~Xu, and D.~Moloney, ``Benchmarking of cnns for
  low-cost , low-power robotics applications,'' in \emph{RSS 2017 Workshop: New
  Frontier for Deep Learning in Robotics}, 2010.

\end{thebibliography}





\end{document}